\documentclass[11pt]{article}
\usepackage{colacl}
\newcommand{\ul}{\underline}

\title{Dialogue Act Tagging with Transformation-Based Learning}
\author{Ken Samuel \and Sandra Carberry \and K. Vijay-Shanker\\
	Department of Computer and Information Sciences\\
	University of Delaware\\
	Newark, Delaware 19716 USA\\
	\{samuel,carberry,vijay\}@cis.udel.edu\\
	http://www.eecis.udel.edu/\~{ }\{samuel,carberry,vijay\}/}

\begin{document}

\maketitle

\begin{abstract}
For the task of recognizing {\em dialogue acts}, we are applying the
Transformation-Based Learning (TBL) machine learning algorithm. To
circumvent a sparse data problem, we extract values of well-motivated
features of utterances, such as speaker direction, punctuation marks,
and a new feature, called {\em dialogue act cues}, which we find to be
more effective than cue phrases and word n-grams in practice. We
present strategies for constructing a set of dialogue act cues
automatically by minimizing the entropy of the distribution of
dialogue acts in a training corpus, filtering out irrelevant dialogue
act cues, and clustering semantically-related words. In addition, to
address limitations of TBL, we introduce a Monte Carlo strategy for
training efficiently and a committee method for computing confidence
measures. These ideas are combined in our working implementation,
which labels held-out data as accurately as any other reported system
for the dialogue act tagging task.
\end{abstract}

\section{Introduction}

Although machine learning approaches have achieved success in many
areas of Natural Language Processing, researchers have only recently
begun to investigate applying machine learning methods to
discourse-level
problems~\cite{Reithinger97,DiEugenio97,Wiebe97,Andernach96,Litman94}.
An important task in discourse understanding is to interpret an
utterance's {\em dialogue act}, which is a concise abstraction of a
speaker's intention, such as SUGGEST and ACCEPT. Recognizing dialogue
acts is critical for discourse-level understanding and can also be
useful for other applications, such as resolving ambiguity in speech
recognition. However, computing dialogue acts is a challenging task,
because often a dialogue act cannot be directly inferred from a
literal interpretation of an utterance.

We have investigated applying Trans\-for\-ma\-tion-Based Learning
(TBL) to the task of computing dialogue acts. This method, which has
not been used previously in discourse, has a number of attractive
characteristics for our task. However, it also has some limitations,
which we address with a Monte Carlo strategy that significantly
improves the training time efficiency without compromising accuracy
and a committee method that enables TBL to compute confidence measures
for the dialogue acts assigned to utterances.

Our machine learning algorithm makes use of abstract features
extracted from utterances. In addition, we utilize an
entropy-minimization approach to automatically identify {\em dialogue
act cues}, which are words and short phrases that serve as signals for
dialogue acts. Our experiments demonstrate that dialogue act cues tend
to be more effective than cue phrases and word n-grams, and this
strategy can be further improved by adding a filtering mechanism and a
semantic-clustering method. Although we still plan to implement more
modifications, our system has already achieved success rates
comparable to the best reported results for computing dialogue acts.

\section{Transformation-Based Learning}

To compute dialogue acts, we are using a modified version of
Brill's~\shortcite{Brill95a} Transformation-Based Learning method.
Given a tagged training corpus, TBL develops a learned model that
consists of a sequence of rules. For example, in one experiment, our
system produced 213 rules; the first five rules are presented in
Figure~\ref{ex-rules}. To label a new corpus of dialogues with
dialogue acts, the rules are applied, in turn, to every utterance in
the corpus, and each utterance that satisfies the conditions of a rule
is relabeled with that rule's new tag. For example, the first rule in
Figure~\ref{ex-rules} labels every utterance with the tag SUGGEST.
Then, after the second, third, and fourth rules are applied, the fifth
rule changes an utterance's tag to REJECT if it includes the word
``no'', and the preceding utterance is currently tagged SUGGEST. Note
that an utterance's tag may change several times as the different
rules in the sequence are applied.

\begin{figure}[ht]
\centering
\begin{tabular}{|c|l|c|}
\hline
\# & \multicolumn{1}{c|}{Condition(s)} & New Tag \\
\hline
1 & {\em none}                    & SUGGEST\\
\hline
2 & Includes ``see'' \& ``you''   & BYE    \\
\hline
3 & Includes ``sounds''           & ACCEPT \\
\hline
4 & Length $<$ 4 words            & GREET  \\
  & Prec. tag is {\em none}\footnotemark &        \\
\hline
5 & Includes ``no''               & REJECT \\
  & Prec. tag is SUGGEST          &        \\
\hline
\end{tabular}
\caption{Rules produced by the system}
\label{ex-rules}
\end{figure}

\footnotetext{This condition is true only for the first utterance of a
dialogue.}

To develop a sequence of rules from a tagged training corpus, TBL
attempts to produce rules that will correctly label many of the
utterances in the training data. The system first generates all of the
{\em potential rules} that would make at least one label in the
training corpus correct. For each potential rule, its {\em improvement
score} is defined to be the number of correct tags in the training
corpus after the rule is applied {\em minus} the number of correct
tags in the training corpus before the rule is applied. The potential
rule with the highest improvement score is applied to the entire
training corpus and output as the next rule in the learned model. This
process repeats (using the new tags assigned to utterances in the
training corpus), producing one rule for each pass through the
training data, until no rule can be found with an improvement score
that surpasses some predefined threshold, $\mathrm{\Theta}$.

Since there are potentially an infinite number of rules that could
produce the dialogue acts in the training data, it is necessary to
restrict the range of patterns that the system can consider by
providing a set of rule templates. The system replaces variables in
the templates with appropriate values to generate rules. For example,
the following template can be instantiated with \ul{w}=``no'',
\ul{X}=SUGGEST, and \ul{Y}=REJECT to produce the last rule in
Figure~\ref{ex-rules}.\\

\noindent
IF utterance \ul{u} contains the word \ul{w}
\\
\noindent
AND the tag on the utterance preceding \ul{u} is \ul{X} \\
\noindent
THEN change \ul{u}'s tag to \ul{Y} \\

We have observed that TBL has a number of attractive characteristics
for the task of computing dialogue acts. TBL has been effective on a
similar\footnote{The part-of-speech tag of a word is dependent on the
word's internal features and on the surrounding words; similarly, the
dialogue act of an utterance is dependent on the utterance's internal
features and on the surrounding utterances.} task, Part-of-Speech
Tagging~\cite{Brill95a}. Also, TBL's rules are relatively intuitive,
so a human can analyze the rules to determine what the system has
learned and perhaps develop a theory. TBL is very good at discarding
irrelevant rules, because the effect of irrelevant rules on a training
corpus is essentially random, resulting in low improvement scores. In
addition, our implementation can accommodate a wide variety of
different types of features, including set-valued features, features
that consider the context of surrounding utterances, and features that
can take distant context into account. These and other attractive
characteristics of TBL are discussed further in \newcite{Samuel98c}.
 
\section{Dialogue Act Tagging}

To address a significant concern in machine learning, called the
sparse data problem, we must select an appropriate set of features.
Researchers in discourse, such as \newcite{Grosz86a},
\newcite{Lambert93}, \newcite{Hirschberg93}, \newcite{Chen95},
\newcite{Andernach96}, \newcite{Samuel96}, and \newcite{Chu-Carroll98}
have suggested several features that might be relevant for the task of
computing dialogue acts. Our system can consider the following
features of an utterance: 1)~the cue phrases\footnote{This feature is
defined later in this section.} in the utterance; 2)~the word
n-grams\footnotemark[3] in the utterance; 3)~the dialogue act
cues\footnotemark[3] in the utterance; 4)~the entire utterance for
one-, two-, or three-word utterances; 5)~speaker
information\footnote{In our system, we are handling speaker
information differently from the previous research. For example,
\newcite{Reithinger97} combine the speaker direction with the dialogue
act to make act-speaker pairs, such as $<$SUGGEST,A$\rightarrow$B$>$
and $<$REJECT,B$\rightarrow$A$>$. But we believe it is more effective
to use the change of speaker feature, which is defined to be false if
the speaker of the current utterance is the same as the speaker of the
immediately preceding utterance, and true otherwise.} for the
utterance; 6)~the punctuation marks found in the utterance; 7)~the
number of words in the utterance; 8)~the dialogue acts on the
preceding utterances; and 9)~the dialogue acts on the
following\footnote{If the system is participating in the dialogue,
rather than simply listening, the future context may not always be
available. But for an utterance that is in the middle of a speaker's
turn, it is reasonable to consider the subsequent utterances within
that same turn. And also, when utterances from the later turns do
become available, it may be important to use this information to
re-evaluate any dialogue acts that were computed and determine if the
system might have misunderstood.} utterances. Other features that we
still plan to implement include: 10)~surface speech acts, to represent
the syntactic structure of the utterance in an abstract format;
11)~the focusing information, specifying which preceding utterance
should be considered the most salient when interpreting the current
utterance; 12)~the type of the subject of the utterance; and 13)~the
type of the main verb of the utterance.

Like other researchers, we recognize that the specific {\em word
substrings} (words and short phrases) in an utterance can provide
important clues for discourse processing, so we should utilize a
feature that captures this information. \newcite{Hirschberg93} and
\newcite{Knott96} have identified sets of {\em cue phrases}.
Unfortunately, we have found that these manually-selected sets of cue
phrases are insufficient for our task, as they were motivated by
different domains and tasks, and these sets may be incomplete.

\newcite{Reithinger97} utilized {\em word n-grams}, which are {\em
all} of the word substrings (with a reasonable bound on the length) in
the training corpus. However, although TBL is capable of discarding
irrelevant rules, if it is bombarded by an overwhelming number of
irrelevant rules, performance may begin to suffer. This is because the
improvement scores of irrelevant rules are random, so if the system
generates too many of these rules, some of their scores might, by
chance, be high enough for selection in the final model, where they
can affect performance on new data.

\begin{figure*}[ht]
\centering
\begin{tabular}{|r|r|c|}
\hline
Category & \# & Examples \\
\hline
\hline
Traditional cues &  56 & ``and'', ``because'', ``but'', ``so'', ``then'' \\
\hline
Potential cues   &  71 & ``bye'', ``how 'bout'', ``see you'', ``sounds'', ``thanks'' \\
\hline
Domain cues      &  42 & ``busy'', ``meet'', ``o'clock'', ``tomorrow'', ``what time'' \\
\hline
Superstring cues & 690 & ``and then'', ``but the'', ``how 'bout the'', ``okay I'', ``so we'' \\
...with filtering  & 472 & ``and then'', ``but the'', ``no I'', ``okay with'', ``so we'' \\
\hline
Undesirable cues         & 170 & ``a'', ``be'', ``had'', ``in the'', ``to'' \\
\hline
\end{tabular}
\caption{A set of dialogue act cues divided into five categories}
\label{cues-categories}
\end{figure*}

As a happy medium between the two extremes of using a small set of
hand-picked cue phrases and considering the complete set of word
n-grams, we are automating the analysis of the training corpus to
determine which word substrings are relevant. We introduce a new
feature called {\em dialogue act cues}: word substrings that appear
frequently in dialogue and provide useful clues to help determine the
appropriate dialogue acts. To collect dialogue act cues automatically
from a training corpus, our strategy is to select word substrings of
one, two, or three words to minimize the entropy of the distribution
of dialogue acts given a substring. A substring is selected if the
dialogue acts co-occurring with it have a sufficiently low entropy,
discarding sparse data. Specifically,

\[\mathrm{C \stackrel{def}{=} \{s \!\in \!S\ |\ H(D|s) \!< \!\theta_1 \wedge \#(s) \!> \!\theta_2\}}\]

\noindent
where C is the set of dialogue act cues, S is the set of word
substrings, D is the set of dialogue acts, $\mathrm{\theta_1}$ and
$\mathrm{\theta_2}$ are predefined thresholds, \#(x) is the number of
times an event, x, occurs in the training corpus, and
entropy\footnote{The entropy is capturing the distribution of dialogue
acts for utterances with a given word substring. By minimizing
entropy, we are selecting a word substring if it produces a highly
skewed distribution of the dialogue acts, and thus, if this word
substring is found in an utterance, it is relatively easy to determine
the proper dialogue act.} is defined in the standard way:\footnote{In
practice, we estimate the probabilities with: \(\mathrm{P(d|s) \approx
\frac{\#(d\&s)}{\#(s)}}\).} \(\mathrm{H(D|s) \stackrel{def}{=} -
\sum_{d \in D} P(d|s) \log_2 P(d|s).}\)

The desirable dialogue act cues produced by our experiments can be
organized into three categories. {\em Traditional cues} are those cue
phrases that have previously been reported in the literature, such as
``but'' and ``so''; {\em potential cues} consist of other useful word
substrings that have not been considered, such as ``thanks'' and ``see
you''; and for dialogues from a particular domain, there may be {\em
domain cues} --- for example, the appointment-scheduling corpora have
dialogue act cues, such as ``what time'' and ``busy''. Dialogue act
cues in the first two categories can be utilized for learning general
rules that should apply across domains, while the third category
constitutes information that can fine-tune a model for a particular
domain.

But this method is not sufficiently restrictive; it selects many word
substrings that do not signal dialogue acts. In many cases, an
undesirable dialogue act cue {\em contains} a useful dialogue act cue
as a substring, so it should be relatively easy to eliminate. Examples
of these {\em superstring cues} include ``but the'' and ``okay I''. We
have implemented a straightforward filtering function to address this
problem. If a dialogue act cue, such as ``how 'bout the'' is subsumed
by a more general dialogue act cue with a better entropy score, such
as ``how 'bout'', then the first dialogue act cue only offers
redundant information, and so it should be removed from the set of
dialogue act cues to minimize the number of irrelevant rules that are
generated. Our filter deletes a dialogue act cue if one of its
substrings happens to be another dialogue act cue with a better or
equivalent entropy score.

Another effective heuristic is to cluster certain dialogue act cues
into semantic classes, which can collapse several potential rules into
a single rule with significantly more data supporting it. For example,
in the appointment-scheduling corpora, there is a strong correlation
between weekdays and the SUGGEST dialogue act, but to express this
fact, it is necessary to generate five separate rules. However, if the
five weekdays are combined under one label, ``\$weekday\$'', then the
same information can be captured by a single rule that has five times
as much data supporting it: ``\$weekday\$'' $\Longrightarrow$ SUGGEST.
We have experimented with clusters, such as ``\$weekday\$'',
``\$month\$'', ``\$number\$'', ``\$ordinal-number\$'', and
``\$proper-name\$''.

We collected a set of dialogue act cues, clustering words in six
semantic classes, with $\mathrm{\theta_1 = H(T)}$ (the entropy of the
dialogue acts) and $\mathrm{\theta_2 = 6}$. As shown in
Figure~\ref{cues-categories}, these dialogue act cues were distributed
among the four categories described above, with an additional category
for the remaining {\em undesirable cues}. Note that our simple
filtering technique successfully eliminated 218 of the superstring
cues. We plan to investigate more sophisticated filtering approaches
to target the remaining 472 superstring cues.

\section{Limitations of TBL}

Although we have argued for the use of Transformation-Based Learning
for dialogue act tagging, we have discovered a significant limitation
of the algorithm: The rule templates used by TBL must be developed by
a human, in advance. Since the omission of any relevant templates
would handicap the system, it is essential that these choices be made
carefully. But, in dialogue act tagging, nobody knows exactly which
features and feature interactions are relevant, so we would prefer to
err on the side of caution by constructing an overly-general set of
templates, allowing the system to {\em learn} which templates are
effective. Unfortunately, in training, TBL must generate {\em all} of
the potential rules for each utterance during each pass through the
training data, and our experimental results indicate that it is
necessary to severely limit the number of potential rules that may be
generated, or the memory and time costs are so exorbitant that the
method becomes intractable.

Our solution to this problem is to implement a Monte Carlo version of
TBL to relax the restriction that TBL must perform an exhaustive
search. In a given pass through the training data, for each utterance
that is incorrectly tagged, only R of the possible template
instantiations are randomly selected, where R is a parameter that is
set in advance. As long as R is large enough, there doesn't appear to
be any significant degradation in performance. We believe that this is
because the best rules tend to be effective for many different
utterances, so there are many opportunities to find these rules during
training; the better a rule is, the more likely it is to be generated.
So, although random sampling will miss many rules, it is still highly
likely to find the best rules.

Experimental tests show that this extension enables the system to
efficiently and effectively consider a large number of potential
rules. This increases the applicability of the TBL method to tasks
where the relevant features and feature interactions are not known in
advance as well as tasks where there are {\em many} relevant features
and feature interactions. In addition, it is no longer critical that
the human developer identify a minimal set of templates, and so this
improvement decreases the labor demands on the human developer.

Unlike probabilistic machine learning approaches, TBL fails to offer
any measure of confidence in the tags that it produces. Confidence
measures are useful in a wide variety of ways; for example, we foresee
that our module for tagging dialogue acts can potentially be
integrated into a larger system so that, when TBL cannot produce a tag
with high confidence, other modules may be invoked to provide more
evidence. Unfortunately, due to the nature of the TBL method,
straightforward approaches for tracking the confidence of a rule
during training have been unsuccessful. To address this problem, we
are using the Committee-Based Sampling method~\cite{Dagan95} and the
Boosting method~\cite{Freund96} in a novel way: The system is trained
multiple times, to produce a few different but reasonable models for
the training data.\footnote{With the efficiencies introduced by our
use of features, dialogue act cue selection, and the Monte Carlo
approach, we can implement modifications that require multiple
executions of the algorithm, which would be infeasible otherwise.} To
construct these models, we adopted the strategy introduced in the
Boosting method, by biasing the later models to focus on those
utterances (in the training set) that the earlier models tagged
incorrectly. Then, given new data, each model independently tags the
input, and the responses are compared. A given tag's confidence
measure is based on how well the different models agree on that tag.
Our preliminary results with five models show that this strategy
produces useful confidence measures --- for nearly half of the
utterances, all five models agreed on the tag, and over 90\% of those
tags were correct. In addition, the overall accuracy of our system
increased significantly. More details on this work are presented in
\newcite{Samuel98c}.

\section{Experimental Results}

A survey of the other research projects that have applied machine
learning methods to the dialogue act tagging task is presented in
\newcite{Samuel98a}. The highest success rate was reported by
\newcite{Reithinger97}, whose system could correctly label 74.7\% of
the utterances in a test corpus. Their work utilized an N-Grams
approach, in which an utterance's dialogue act was based on substrings
of words as well as the dialogue acts and speaker information from the
preceding two utterances. Various probabilities were estimated from a
training corpus by counting the frequencies of specific events, such
as the number of times that each pair of consecutive words co-occurred
with each dialogue act.

As a direct comparison, we applied our system to Reithinger and
Klesen's training set (143 dialogues, 2701 utterances) and disjoint
testing set (20 dialogues, 328 utterances), which consist of
utterances labeled with 18 different dialogue acts. Using semantic
clustering, $\mathrm{\Theta = 1}$ (the improvement score threshold),
$\mathrm{R = 14}$ (the Monte Carlo sample size), a set of dialogue act
cues, change of speaker, the dialogue act on the preceding utterance,
and other features, our system achieved an average accuracy score over
five\footnote{This is to factor out the random aspect of the Monte
Carlo method.} runs of 75.12\% ($\sigma$=1.34\%), including a high
score of 77.44\%. We have also run direct comparisons between our
system and Decision Trees, determining that our system's performance
is also comparable to this popular machine learning
method~\cite{Samuel98c}.

Figure~\ref{cues-results} presents a series of experiments which vary
the set of word substrings utilized by the system.\footnote{Note that
these results cannot be compared with the results presented above,
since several parameter values differ between the two sets of
experiments.} Each experiment was run ten times, and the results were
compared using a two-tailed t test to determine that all of the
accuracy differences were significant at the 0.05 level, except for
the differences between rows 3 \& 4, rows 4 \& 5, rows 4 \& 6, rows 5
\& 6, rows 5 \& 7, and rows 6 \& 7.

\begin{figure*}[ht]
\centering
\begin{tabular}{r|r||c}
Word Substrings & \# & Accuracy \\
\hline
None                                                &    0 & 41.16\% ($\sigma$=0.00\%) \\
Cue phrases (from previous literature)\footnotemark &  936 & 61.74\% ($\sigma$=0.69\%) \\
Word n-grams                                        &16271 & 69.21\% ($\sigma$=0.94\%) \\
Entropy minimization                                & 1053 & 69.54\% ($\sigma$=1.97\%) \\
Entropy minimization with clustering                & 1029 & 70.18\% ($\sigma$=0.75\%) \\
Entropy minimization with filtering                 &  826 & 70.70\% ($\sigma$=1.31\%) \\
Entropy minimization with filtering and clustering  &  811 & 71.22\% ($\sigma$=1.25\%) \\
\end{tabular}
\caption{Tagging accuracy on held-out data, using different sets of
word substrings in training}
\label{cues-results}
\end{figure*}

\footnotetext{There are only 478 different cue phrases in the set, but
for our system, it was necessary to manipulate the data in various
ways, such as including a capitalized version of each cue phrase and
splitting up contractions.}

As the figure shows, when the system was restricted from using any
word substrings, its accuracy on unseen data was only 41.16\%. When
given access to all of the cue phrases proposed in previous
work,\footnote{See Hirschberg and Litman~(1993) and Knott~(1996) for
these lists of cue phrases. We also included 45 cue phrases that we
pinpointed by manually analyzing a completely different set of
dialogues, two years before we began working with the \scshape
VerbMobil \normalfont corpora.} the accuracy rises
significantly~($\mathrm{p<0.001}$) to 61.74\%. But this result is
significantly lower~($\mathrm{p<0.001}$) than the 69.21\% accuracy
produced by using all substrings of one, two, or three words (word
n-grams) in the training data, as \newcite{Reithinger97} did. And the
entropy-minimization approach with the filtering and clustering
techniques produce dialogue act cues that cause the accuracy to rise
significantly further~($\mathrm{p=0.003}$) to 71.22\%.

Our experimental results show that the cue phrases identified in the
literature do not capture all of the word substrings that signal
dialogue acts. On the other hand, the complete set of word n-grams
causes the performance of TBL to suffer. Our dialogue act cues
generate the highest accuracy scores, using significantly fewer word
substrings than the word n-grams approach.

\section{Discussion}

This paper has presented the first attempt to apply
Transformation-Based Learning to discourse-level problems. We utilized
various features of utterances to learn effectively from a relatively
small amount of data, and we have developed an entropy-minimization
approach with filtering and clustering that automatically collects
useful dialogue act cues from tagged training data. In addition, we
have devised a Monte Carlo strategy and a committee method to address
some limitations of TBL. Although we have only begun implementing our
ideas, our system has already matched Reithinger and Klesen's success
rate in computing dialogue acts.

In the future, we plan to implement more features, improve our method
for collecting dialogue act cues, and investigate how these
modifications improve our system's performance. Also, for the
semantic-clustering technique, we selected the clusters of words by
hand, but it would be interesting to see how a taxonomy, such as
WordNet could be used to automate this process.

When there is not enough tagged training data available, we would like
the system to learn from untagged data. Dagan and
Engelson's~\shortcite{Dagan95} Committee-Based Sampling method
constructed multiple learned models from a small set of tagged data,
and then, only when the models disagreed on a tag, a human was
consulted for the correct tag. \newcite{Brill95b} developed an
unsupervised version of TBL for Part-of-Speech Tagging, but this
algorithm must be initialized with words that can be tagged
unambiguously,\footnote{For example, ``the'' is always a Determiner.}
and in discourse, there are very few unambiguous examples. We intend
to investigate a weakly-supervised approach that utilizes the
confidence measures described above. First, the system will be trained
on a relatively small set of tagged data, producing a few different
models. Then, given untagged data, it will use the models to derive
dialogue acts with confidence measures. Those tags that receive high
confidence can be used as unambiguous examples to drive the
unsupervised version of TBL.

While we contend that machine learning can be an effective tool for
identifying dialogue acts, we do realize that machine learning may not
be able to completely solve this problem, as it is unable to capture
some relevant factors, such as common-sense {\em world knowledge}. We
envision that our system may potentially be integrated into a larger
system that uses confidence measures to determine when world knowledge
information is required.

\section{Acknowledgments}

We wish to thank the members of the \scshape VerbMobil \normalfont
research group at DFKI in Germany, particularly Norbert Reithinger,
Jan Alexandersson, and Elisabeth Maier, for providing us with the
opportunity to work with them and generously granting us access to the
\scshape VerbMobil \normalfont corpora. This work was partially
supported by the NSF Grant \#GER-9354869.

\bibliographystyle{acl}
\bibliography{/usa/samuel/class/research/related_research/mybibfile}

\end{document}